\documentclass[11pt,a4paper]{article}
\usepackage[hyperref]{acl2020}
\usepackage{times}
\usepackage{latexsym}

\usepackage{microtype}

\aclfinalcopy 
\def\aclpaperid{2431} 

\newcommand{\exper}[1]{\textsc{#1}}
\newcommand{\Ours}{\exper{PC}}

\definecolor{restt}{RGB}{206, 1, 96}
\definecolor{eatt}{RGB}{250, 100, 14}
\definecolor{pricee}{RGB}{112, 54, 42}
\definecolor{ratee}{RGB}{22, 113, 207}
\definecolor{nearr}{RGB}{21, 62, 164}
\definecolor{foodd}{RGB}{80, 179, 66}
\definecolor{areaa}{RGB}{183, 122, 41}
\definecolor{otherr}{RGB}{1,1,1}

\newcommand{\restt}[1]{{{\color{restt}#1}}}
\newcommand{\eatt}[1]{{{\color{eatt}#1}}}
\newcommand{\pricee}[1]{{{\color{pricee}#1}}}
\newcommand{\ratee}[1]{{{\color{ratee}#1}}}
\newcommand{\nearr}[1]{{{\color{nearr}#1}}}
\newcommand{\foodd}[1]{{{\color{foodd}#1}}}
\newcommand{\areaa}[1]{{{\color{areaa}#1}}}
\newcommand{\otherr}[1]{{\text{\color{otherr}#1}}}

\title{Posterior Control of Blackbox Generation}

\author{Xiang Lisa Li \\
  Department of Computer Science\\
  Johns Hopkins University \\
  \texttt{xli150@jhu.edu} \\\And
  Alexander M. Rush \\
  Department of Computer Science \\
  Cornell Tech \\
  \texttt{arush@cornell.edu} \\}

\date{}

\usepackage{amsmath,amsfonts,bm}

\def\eqref#1{equation~\ref{#1}}

\def\1{\bm{1}}

\DeclareMathAlphabet{\mathsfit}{\encodingdefault}{\sfdefault}{m}{sl}
\SetMathAlphabet{\mathsfit}{bold}{\encodingdefault}{\sfdefault}{bx}{n}

\newcommand{\softmax}{\mathrm{softmax}}

\DeclareMathOperator*{\argmax}{arg\,max}

\usepackage{hyperref}
\usepackage{url}
\usepackage{wrapfig}

\usepackage{times}
\usepackage{latexsym}
\usepackage{microtype}
\usepackage{natbib}
\usepackage{booktabs}
\usepackage{appendix}
\usepackage{tikz}
\usepackage{tikz-dependency}
\usepackage{amsmath}
\usepackage{amsfonts}
\usepackage{graphicx}
\usepackage{amsthm}
\usepackage{mathtools}
\usepackage{multirow}
\usepackage{url}
\usepackage{color}
\usepackage{subfig}
\usepackage{dsfont}
\usepackage[ruled,vlined,noend]{algorithm2e}

 \usepackage[disable]{todonotes}
\makeatletter
\newcommand*\iftodonotes{\if@todonotes@disabled\expandafter\@secondoftwo\else\expandafter\@firstoftwo\fi}  
\makeatother

\makeatletter
\newcommand\footnoteref[1]{\protected@xdef\@thefnmark{\ref{#1}}\@footnotemark}
\makeatother

\newlength{\extramargin}
\usepackage{calc}
\iftodonotes{\usepackage[paperwidth=\paperwidth+\extramargin*2,marginparwidth=\marginparwidth+\extramargin,width=\textwidth,height=\textheight]{geometry}}{}

\definecolor{oracle}{rgb}{0.5, 0.5, 0.5}

\usepackage[noabbrev,capitalize]{cleveref} 

\crefname{equation}{equation}{equations}   
\crefname{footnote}{footnote}{footnotes}   
\crefname{line}{line}{lines}               
\crefname{section}{\S}{\S\S}
\Crefname{section}{\S}{\S\S}    
\crefformat{section}{#2\S#1#3}  
\Crefformat{section}{#2\S#1#3}
\crefrangeformat{section}{\S\S#3#1#4--#5#2#6}
\Crefrangeformat{section}{\S\S#3#1#4--#5#2#6}

\DeclareMathOperator{\KL}{KL}

\DeclareMathOperator{\Entr}{H}

\newcommand{\ptheta}{p_\theta}
\newcommand{\qphi}{q_\phi}

\DeclareCaptionLabelFormat{andtable}{#1~#2  \&  \tablename~\thetable}

\usepackage{lipsum}

\DeclareMathOperator{\Softmax}{softmax}
\DeclareMathOperator{\score}{score}
\DeclareMathOperator{\logsumexp}{logsumexp}

\begin{document}
\maketitle
\begin{abstract}

Text generation often requires high-precision output that obeys task-specific rules. This fine-grained control is difficult to enforce with off-the-shelf deep learning models. In this work, we consider augmenting neural generation models with discrete control states learned through a structured latent-variable approach. Under this formulation, task-specific knowledge can be encoded through a range of rich, posterior constraints that are effectively trained into the model. 
This approach allows users to ground internal model decisions based 
on prior knowledge, without sacrificing the representational power of neural generative models. Experiments consider applications of this approach for text generation. We find that this method improves over standard benchmarks, while also providing fine-grained control. 

\end{abstract}

\section{Introduction}

A core challenge in using deep learning for NLP is developing methods that allow for controlled output while maintaining the broad coverage of data-driven methods. While this issue is less problematic in classification tasks, it has hampered the deployment of systems for conditional natural language generation (NLG), where users often need to control output through task-specific knowledge or plans. While there have been significant improvements in generation quality from automatic systems \citep{Mei_2016,Du_ek_2016,Lebret_2016}, these methods are still far from being able to produce controlled output \citep{Wiseman_2017}. Recent state-of-the-art system have even begun to utilize manual control through rule-based planning modules \cite{moryossef-etal-2019-step, puduppully-etal-2019-data}.

Consider the case of encoder-decoder models for generation, built with RNNs or transformers. These models generate fluent output and provide flexible representations of their conditioning. Unfortunately, auto-regressive decoders are also globally dependent, which makes it challenging to 
incorporate domain constraints. 

Research into controllable deep models aims to circumvent the all-or-nothing dependency trade-off of encoder-decoder systems and expose explicit higher-level decisions.
 One line of research has looked at global control states that represent sentence-level properties for the full decoder. For example, \citet{hu2017controlled} uses generative adversarial networks where the attributes of the text (e.g., sentiment, tense) are exposed. Another line of research exposes fine-level properties, such as phrase type, but requires factoring the decoder to expose local decisions, e.g.  \citet{template-generation}.

This work proposes a method for augmenting any neural decoder architecture to incorporate fine-grained control states. The approach first modifies training to incorporate structured latent control variables. Then, training constraints are added to anchor the state values to problem-specific knowledge. At test time, the control states can be ignored or 
utilized as grounding for test-time constraints.
Technically, the approach builds on recent advances in structured amortized variational inference to enforce additional constraints on the learned distribution. These constraints are enforced through efficient structured posterior calculations and do not hamper modeling power.

We demonstrate that the method can improve accuracy and control, while utilizing a range of different posterior constraints. In particular on two large-scale data-to-text generation datasets, E2E \citep{e2edata} and WikiBio \citep{wikibiodata}, our method increases the performance of benchmark systems while also producing outputs that respect the grounded control states. Our code is available at \url{https://github.com/XiangLi1999/PosteriorControl-NLG}.

\begin{figure*}
    \centering
    \includegraphics[page=1,width=1\textwidth]{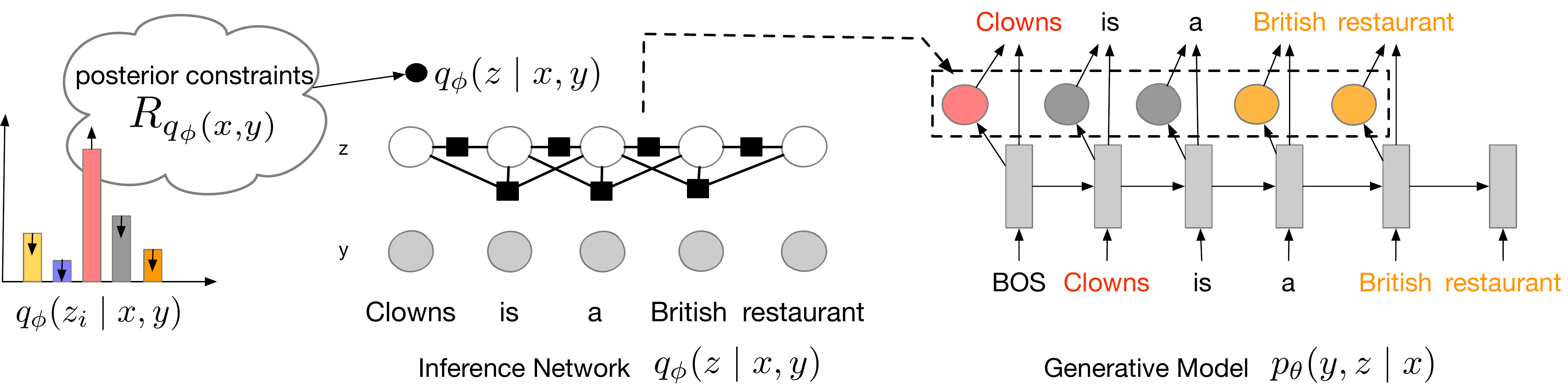}
    \caption{\label{fig:data} Model training. Assumes we are given conditioning $x$ (not shown) and output sentence $y$. (Middle) An inference network $\phi$ is used to parameterize a structured segmental conditional random field $q_\phi(z \mid x, y)$ over control states $z$. (Right) Sample from $q_\phi$ (colored circles) is used to provide control state labels for a blackbox generation model $p_{\theta}(y, z\mid x)$ . (Left) To ground the control states to represent problem-specific meaning, posterior regularization is used to enforce distributional constraints through penalties $R_q(x,y)$. The whole system is optimized end-to-end to learn latent properties of the final output tokens. }
    \vspace{-7pt}
\end{figure*}

\section{Control States for Blackbox Generation}

Consider a conditional generation setting where the input consists of an arbitrary context $x$ and the output $y_{1:T}$ is a sequence of target tokens. We are interested in modeling latent fine-grained, discrete control states $z = z_{1:T}$ each with a label in $\cal C$. We assume that these states are weakly-supervised at training through problem-specific constraints. The goal is to induce a model of $p(y \mid x) = \sum_z p(y, z \mid x)$. Concretely, our experiments will focus on a data-to-text generation problem where $x$ corresponds to a table of data, and $y_{1:T}$ is a textual description. We hope to induce control states $z$ that indicate which table fields are being described, and our weak supervision corresponds to indicators of known alignments.

We assume the generative model is a blackbox auto-regressive decoder that produces both $y$ and $z$.  Define this general model as:
\begin{eqnarray*}   
\ptheta(y, z \mid x ) = &  \prod_{t=1}^T & p_\theta(y_t \mid x, y_{< t}, z_{\leq t}) \cdot \\
&& p_\theta(z_t \mid x, y_{< t}, z_{< t})
\end{eqnarray*}
For a neural decoder, where $h_t(y_{1:t-1}, z_{1:t-1})$ is the hidden state at time-step $t$, we might generate the latent class $z_t \in \cal C$ and next token $y_t$ as,
\begin{eqnarray*}
 \ptheta(z_t \mid z_{< t}, y_{< t}) \hspace*{-0.2cm} &=& \hspace*{-0.2cm}\Softmax(W_0  h_t + b_0) \\
\ptheta(y_t \mid z_{\leq t}, y_{< t}) \hspace*{-0.2cm} &=& \hspace*{-0.2cm} \Softmax(W_1  [h_t, g_{\theta}(z_t)] + b_1)
\end{eqnarray*}
\noindent
Here $g_{\theta}$ is a parameterized embedding function and $W, b$ are model parameters from $\theta$. The log-likelihood of the model is given by
$\mathcal{L}(\theta) =  \log p_\theta(y \mid x)$.

The key latent term of interest is the posterior distribution $p_\theta(z \mid x, y)$, i.e. the probability of over state sequences for a known output. The decoder parameterization makes this distribution intractable to compute in general. We instead use variational inference to define a parameterized variational posterior distribution, $\qphi(z\mid x,y)$, from a preselected family of possible distributions $\cal Q$.\footnote{Since our family is over a combinatorial set of $z_{1:T}$, this corresponds to a \textit{structured} variational inference setting.} To fit the model parameters $\theta$, we utilize the evidence lower bound (for any variational parameters $\phi$),
\begin{align*}
&{\cal L}(\theta) \geq  \text{ELBO}(\theta, \phi)  \nonumber\\ 
& = \mathbb{E}_{z \sim q_{\phi}(z\mid x,y)}[\log \ptheta( y, z \mid x)]
 + \Entr[q_\phi(z\mid x,y)]
 \label{eq:obj}
\end{align*}

Several recent works have shown methods for effectively fitting neural models with structured  variational inference
\citep{NIPS2016_Johnson,Krishnan:2017,URNNG}. We therefore use these techniques as a backbone for enforcing problem-specific control. See \cref{sec:seg} for a full description of the variational family used.

\section{Posterior Regularization of Control States}

Posterior regularization (PR) is an approach for enforcing soft constraints on the posterior distribution of generative models \citep{ganchev}. Our goal is to utilize these soft constraints to
enforce problem specific weak supervision. 
Traditionally PR uses linear constraints which in the special case of expectation maximization for exponential families leads to convenient closed-form training updates.
As this method does not apply to neural generative models, we resort to gradient-based methods. In this section, we develop a form of posterior regularization that accommodates the neural variational setting.

Starting with the log-likelihood objective, $\mathcal{L}(\theta)$, PR aims to add distributional constraints on the posterior. These soft constraints are expressed as a distributional penalty, $R_p(x, y) \geq 0$.  For example, if we have partial information that a specific control state takes on label $c$ we can add a constraint $R_p(x, y) = 1- p(z_t = c\ |\ x, y)$. We might also consider other distributional properties, for instance penalizing the entropy of a specific posterior marginal, $R_p(x, y) = \Entr_{z'}(z_t=z'\ |\ x, y)$.
See \cref{sec:prNLG} for more constraint examples.

PR uses these soft constraints to regularize the model. Ideally we would penalize the posterior directly, but as noted above, computing this term in a blackbox model is intractable. 
We therefore follow \newcite{ganchev} and use a relaxed version with a surrogate posterior $q_\phi(z \mid x, y)$,
\begin{eqnarray}
\mathcal{L}_{PR}(\theta) &=&  \mathcal{L}(\theta) - \\
 && \hspace{-1cm} \min_{\phi} [\KL[q_{\phi}\ ||\ \ptheta (z\mid x,y)] + \lambda R_{q_{\phi}}(x,y)] \nonumber
\end{eqnarray}
 We can write this in terms of a variational lower-bound on the relaxed PR objective.
\begin{eqnarray}
 \mathcal{L}_{PR}(\theta) &\geq& \text{PRLBO}(\theta, \phi) = \mathcal{L}(\theta) - \\
  && \hspace{-1cm}  [\KL[q_{\phi}\ ||\ \ptheta (z\mid x,y)] + \lambda R_{q_{\phi}}(x,y)] \nonumber
 \end{eqnarray}
  This allows us to relate the $q$ in the PRLBO to the variational posterior in the ELBO simply by expanding the KL and rearranging terms,
\begin{eqnarray*}
  \text{PRLBO}(\theta, \phi) &=& \text{ELBO}(\theta, \phi) -  \lambda R_{q_\phi}(x,y)
\end{eqnarray*}
To train, we jointly maximize over both terms in the PRLBO: the model parameters $\theta$ and the variational parameters $\phi$ (which tightens the bounds). Following standard practice, we use an amortized inference network, i.e. a variational autoencoder \cite{KingmaW13,MnihG14,rezende2014stochastic}, to define $\phi$. 
 
\section{Structured Variational Family for Segmental Generation}
\label{sec:seg}

We now discuss how to efficiently compute the PRLBO under a structured variational family. 
\[ \text{PRLBO} = \underbrace{\mathbb{E}_{z \sim q_{\phi}}[\log \ptheta]}_{(1)}
 + \underbrace{\Entr[q_\phi]}_{(2)} -  \underbrace{\lambda R_{q_\phi}(x,y)}_{(3)} \]
We need a $q_\phi (z \mid x, y)$ for which we can efficiently (1) take samples, (2) compute entropy, and (3) compute the distributional penalties. This motivates the use of a factored conditional random field (CRF), defined by a potential function $\phi (x,y,z)$. At training time, $x,y$ are observed and $z$ is the latent variable that denotes the control states. We then specify a variational posterior distribution: $ q_{\phi}(z \mid x, y) = \frac{ \phi(x, y, z) }{ \sum_{z'}  \phi(x, y, z')}$.

In this work, we focus on the semi-Markov CRF \citep{Young-segmantal-HMM, Sarawagi-SM-CRF}, a common CRF family used in generation \cite{template-generation}. It divides tokens into segmental spans, which are useful for generating entity mentions and commonly used phrases. This model divides the potential function into three parts: the \textbf{emission} potential for a span of tokens given a state, denoted as $\phi_{(e)}$; the \textbf{transition} potential between states, $\phi_{(t)}$; and the \textbf{length} potential of span length given a state, $\phi_{(l)}$. Suppose our control states define a span from $i$ (inclusive) to $j$ (exclusive) labeled by $c$, we denote it as $z_{i:j} = c$. The potential function of a labeled sequence is defined: 
\begin{eqnarray}
\phi (x,y,z) = \prod_{i< j<k} \phi_{(t)}(z_{i:j} , z_{j:k}) \cdot  \phi_{(l)}( j-i) \cdot \nonumber \\
 \phi_{(e)}(x, y_{i:j}, z_{i:j})  
\end{eqnarray}

\begin{algorithm}[t]
       Given $\phi$ and generic semiring $(\oplus,\otimes, \mathbf{0}, \mathds{1})$ \\
       Set $\beta_{T}(c) = \mathds{1}$ $\forall c\in {\cal C}$ \\
    \For{$i = T-1, \ldots, 0$}{
    \For {$c \in {\cal C}$} {
        $\beta'_i(c) = \displaystyle \bigoplus_{d=1}^{\min(L, T-i)} \beta_{i+d}(c) \otimes \phi_{(l)}(d) \otimes   
        \hspace*{2.5cm} \phi_{(e)}(x, y_{i, i+d}, c)$
    }
    \For {$c \in {\cal C}$} {
    $\beta_i(c) = \displaystyle \bigoplus_{c'  \in {\cal C}}^{} \beta'_{i}(c') \otimes \phi_{(t)}(c,c')$
    }
    }
    \Return $Z =\displaystyle \bigoplus_{c \in {\cal C}}^{} \beta'_{0}(c) \otimes \phi_{(t)}(0,c)$
 	\caption{\label{fig:algo} \small Generic Semi-Markov Algorithm.}
\end{algorithm}

\begin{table*}
\small
\hspace*{-0.5cm}\begin{tabular}{l p{6cm} l p{6.3cm}}
\toprule
    \textbf{One-to-One}  & & \textbf{One-to-Many} & \\
    Name & Penalty   &  Name & Penalty \\ 
    \midrule
    Inclusion &    $\text{For}\  (i, j, f) \in A(x, y),$ & 
    Sparsity & For $f \in \mathcal{F}$, \\ 
        & $ \ \  \ R_q =  1 - q( z_{i:j} = \sigma(f) \mid x, y)$  & 
        & \ \ \ $R_q = \ \Entr[\sigma(c \mid f) ]\  $\\
    Exclusion & $\text{For  } f \in x \text{\ and \ } (i, j, f) \not \in A(x, y),$  &
    Fit &  For $(i,j,f) \in A(x,y)$ \\
     & $  \ \ \  R_q = q(z_{i:j} = \sigma(f) \mid x, y)   $ &
     & $\ \ \  R_q = \Entr[\sigma(c \mid f), q(z_{i:j} \mid x, y)]$ \\
    Coverage &   $ \text{For } f \in {\cal F} $, &
    Diversity & Let $p_\text{agg}(\hat{z}) \propto \sum_{t=1}^T q(z_t = \hat{z}\mid x, y)$  \\
    & \ \ \     $\displaystyle R_q = |\sum_{i<j} q(z_{i:j} = \sigma(f)  \mid x, y)- \mathds{1}(f \in x)|$ & & \ \ \ $R_q = 
\Entr[\text{Unif} (\hat{z})] - \Entr[p_\text{agg}(\hat{z})]$ \\
    \bottomrule
\end{tabular}
\caption{\label{tb:pr-constraint}Posterior penalties utilized in the One-to-One and One-to-Many setting. These constraints softly enforce an alignment between control states and text spans by penalizing posterior violations. The objective sums over the three $R_q$ in both cases. }
\end{table*}

\noindent For computational efficiency, we restrict all segment length to be $\leq L$.\footnote{ The time complexity to compute the posterior moments of the full semi-Markov CRF is $O(|{\cal C}|^2nL)$.}

With this model, we can use the forward-backward algorithm for all required inferences: exact sampling, computing partition function, entropy, and posterior marginals $q_{\phi}(z_{i:j} = c\mid x, y)$, useful for term (3). 
In \cref{fig:algo}, we give a generic semi-Markov algorithm \citep{Sarawagi-SM-CRF}.  
We store two tables $\beta$ and $\beta'$, both of size $T \times |{\cal C}|$. 
$\beta_t(c)$ denotes the event that there is a transition at time $t$ from state $c$. $\beta'_t(c)$ denotes the event that there is a emission starting from time $t$ at state $c$. Then we have the recursion for $\beta'_t(c)$ by ``summing'' over different span length, and we have the recursion for $\beta_t(c)$ that sums over all different state transitions.

The algorithm is generic in the sense that different $(\otimes, \oplus)$ operators allow us to compute different needed terms. For example, computing the partition function $Z = \sum_{z'} \phi(x, y, z')$ requires the ($+$,$\times$) semiring \citep{Goodman:semiring-1999,li-eisner-2009}, other distributional terms can be computed by using the same algorithm with alternative semirings and backpropagation \footnote{We need four terms: (a) log-partition term $\log \sum_{z'} \phi(x, y, z')$ requires the log semiring $(\logsumexp, +)$.  The posterior marginals $q(z \mid x, y)$ requires backpropagating from the log-partition term; (b)  max score $\max_z \phi(x,y,z)$: $(\max, +)$ max semiring  and argmax $\argmax_z \phi(x,y,z)$ by (subgradient) backpropagation, (c) entropy through an expectation semiring   $\langle p_1, r_1 \rangle \otimes \langle p_2, r_2 \rangle = \langle p_1 p_2, p_1r_2 + p_2 r_1 \rangle $, and $\langle p_1, r_1 \rangle \oplus \langle p_2, r_2 \rangle = \langle p_1 + p_2, r_1 + r_2 \rangle $, with $\mathds{1} = \langle 1, 0 \rangle$. To initialize, all the emission, transition and length scores takes the form $\langle \phi, -\log \phi \rangle$. The algorithm returns $\langle Z, R \rangle$, and the true entropy is $\frac{R}{Z} + \log Z$. (d) exact sampling through one backward pass and one forward filtering backward sampling, where forward uses the log-partition semiring and backpropagation is by categorical sampling.}. 
\label{ssec:qmodel}

\section{Posterior Constraints from Data Alignment}
\label{sec:prNLG}
\begin{table}
    \begin{tabular}{lp{6cm}}
    \toprule
    $x$ & name[Clowns] eatType[coffee shop],  
    rating[1 out of 5],  near[Clare Hall]\\
    $f \in x$ & \restt{name}, \foodd{eatType},  \ratee{rating}, \nearr{near} \\
    \midrule
     $y$ & \restt{Clowns}$_1$ is$_2$ a$_3$ \foodd{coffee$_4$ shop$_5$} near$_6$ \nearr{Clare$_7$ Hall$_8$} with$_9$ a$_{10}$ \ratee{1$_{11}$ out$_{12}$ of$_{13}$ 5$_{14}$} rating$_{15}$  \\
     $A(x, y)$ & \restt{(1, 2, name)}, \foodd{(4, 6, eatType)}, \nearr{(7, 9, near)}, \ratee{(11, 15, rating)} \\
    \bottomrule
    \end{tabular}
    \caption{\label{tb:example1}Example of data alignment notation. Here $x$ is a table of data, and $f$ are its fields. For a given output $y$ we enforce a soft alignment $A$.}
\end{table}

To make the PR model concrete, we consider the problem of incorporating weak supervision from heuristic alignment in a data-to-text generation task.  Assume that we are tasked with describing a table $x$ consisting of global field names $\cal F$ each with a text value $v$, e.g. $x_f = v$. Not all global fields may be used in a given $x$, we use $f \in x$ to indicate an active field.

We would like control states to indicate when each field is used in generation. Our alignment heuristic is that often these fields will be expressed using the identical text as in the table. While this heuristic obviously does not account for all cases, it is very common in natural language generation tasks as evidence by the wide use of copy attention based approaches \cite{gu-etal-2016-incorporating, gulcehre-etal-2016-pointing}. To utilize these alignments, we use the notation $(i, j, f) \in A(x, y)$ to indicate that a span $i:j$ in the training text $y$ overlaps directly with a field $f \in x$. Table~\ref{tb:example1} gives an example of the notation.

\paragraph{One-to-One Constraints}

\noindent
We first consider one-to-one constraints where we assume that we have a static, mapping from fields to states $\sigma: {\cal F} \mapsto {\cal C}$. Given this mapping, we need to add penalties to encourage the semi-Markov model to overlap with the given weak supervision.   

To enforce soft alignments, we define three posterior constraint types and their computation as shown in Table~\ref{tb:pr-constraint} (Left). The three constraints are i) Inclusion: if a span in $y$ aligns with a field value $f$, then label that span $\sigma(f)$ the state allocated to that field; ii) Exclusion: A span should only have a state $\sigma(f)$, if it aligns with the field value of type $f$; iii) Coverage. The usage count of state $\sigma(f)$ should be $1$ if $f$ in $x$.

\paragraph{One-to-Many Constraints}

We also consider the case when it is infeasible to specify a hard mapping $\sigma$ between the fields and the states. For example, $\cal F$ could be unbounded or large, whereas we hope to keep the cardinality of states small for computational efficiency. 

We propose a method of inducing a dynamic soft mapping $\sigma(c \mid f)$ as we train the model, and impose constraints on the mapping from table field to the state names. First, we would like the distribution of state given table field to be consistent, so one table field is mapped to roughly 1 state. Second, we want to make use of the state space as much as possible by requiring a diverse usage of states.

In order to enforce these properties we introduce the dynamic mapping as a second amortized variational distribution $\sigma(c \mid f; M) = \softmax(Mf)$ which gives the probability that a table field $f$ takes on state $c$.  As shown in Table~\ref{tb:pr-constraint} (Right), we define three constraints that regularize the local $q$ with respect to the global $\sigma$: 
i) Sparsity: Each vocabulary entry in $\sigma$ should have low entropy;
ii) Fit: The global $\sigma$ should represent the class name distribution posterior of each table field by minimizing the cross entropy between types $\sigma(c \mid f )$ and tokens $q(z_{i:j} \mid x, y)$ for all $(i,j,f) \in A(x, y)$;
iii) Diversity: the aggregate class label distribution over all the token in a sentence should have high entropy.

\section{Related Work}

In addition to previously mentioned work, other researchers have noted the lack of control of deep neural networks and proposed methods at sentence-level, word-level, and phrase-level. For example \citet{peng-etal-2018-towards} and \citet{luo-etal-2019-learning} control the sentiment in longer-form story generation. Others aim for sentence-level properties such as sentiment, style, tense, and specificity in generative neural models \citep{hu2017controlled, oraby-2018-controlling, zhang-2018-specifity, shen-2017-style}.  
Closest to this work is that of \citet{template-generation} who control phrase-level content by using a neuralized hidden semi-Markov model for generation itself. Our work differs in that it makes no independence assumption on the decoder model, uses a faster training algorithm, and proposes a specific method for adding constraints.
Finally, there is a line of work that manipulates the syntactic structure of generated texts, by using some labeled syntactic attribute (e.g., parses) or an exemplar \citep{deriu-cieliebak-2018-syntactic-manip, colin-gardent-2018-generating-syntax-para, iyyer-etal-2018-adversarial, chen-etal-2019-controllable}. 
While our work uses control states, there is no inherent assumption of compositional syntax or grammar.

Posterior regularization (PR) is mostly used in standard EM settings to impose constraints on the posterior distribution that would otherwise be intractable (or computationally hard) in the prior. \citet{ganchev} applies posterior regularization to word alignment, dependency parsing, and part-of-speech tagging. Combining powerful deep neural networks with structured knowledge has been a popular area of study: \citet{xu2019multi} applies PR to multi-object generation to limit object overlap; \citet{weakly-supervised} focuses on object detection, and uses PR features to exploit mutual exclusion. In natural language processing; \citet{hu-etal-2016-harnessing,hu-etal-2016-deep} propose an iterative distillation procedure that transfers logic rules into the weights of neural networks, as a regularization to improve accuracy and interpretability. 

Finally, the core of this work is the use of amortized inference/variation autoencoder to approximate variational posterior \citep{KingmaW13,MnihG14,rezende2014stochastic}. We rely heavily on a structure distribution, either linear chain or semi-Markov, which was introduced as a structured VAEs \citep{NIPS2016_Johnson,Krishnan:2017,AmmarDS14}. Our setting and optimization are based on \citet{URNNG}, who introduce a latent tree variable in a variational autoencoding model with a CRF as the inference network,  and on \citet{Yin2018} who use an encoder-decoder model as the inference network.

\begin{figure*}
\begin{minipage}{.4\textwidth}
  \centering
  \small
\resizebox{1\textwidth}{!}{
    \begin{tabular}{l}
    \toprule
    Table ($x$): name[Clowns] eatType[coffee shop]\\ food[Chinese] customer-rating[1 out of 5]
    \\ area[riverside] near[Clare Hall]\\
    \midrule
     \textbf{Ref.1}: Clowns is a coffee shop in the riverside area \\ near Clare Hall that has a  rating 1 out of 5 . \\  They  serve Chinese food .  \\
     \textbf{Ref.2}: The Chinese coffee shop by the riverside near \\ Clare Hall that only has a customer rating of \\ 1  out of 5 is called Clowns .   \\
     \textbf{Ref.3}: There is a Chinese coffee shop near Clare Hall \\  in the riverside area called Clowns its not got \\ a good rating though . \\
    \bottomrule
    \end{tabular}}
\end{minipage}
\begin{minipage}{.6\textwidth}
  \centering
    \begin{tabular}{cp{3.5cm}}
         \raisebox{-3.5cm}{\includegraphics[width=0.4\textwidth]{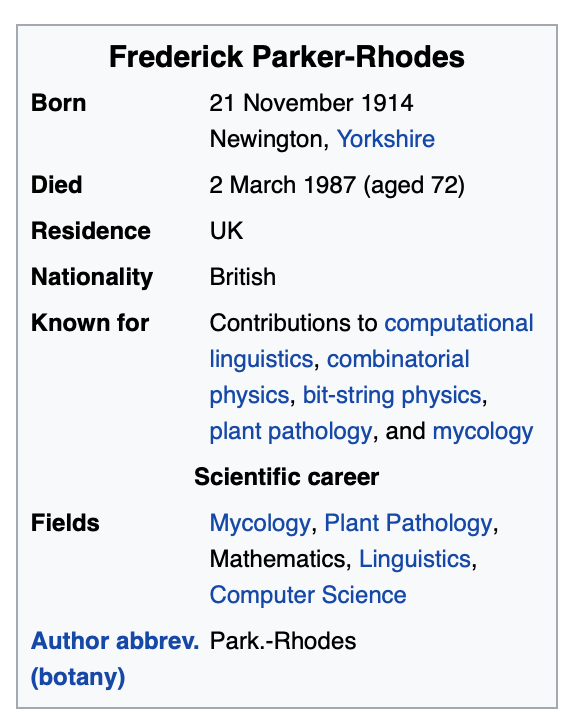}}&  {\small
         \textbf{Ref.1}: Frederick Parker-Rhodes (21 March 1914 -- 21 November 1987) was an English linguist, plant pathologist, computer scientist, mathematician, mystic, and mycologist.}\\
    \end{tabular}
\end{minipage}
\caption{ \label{fig:my_label} Generation benchmarks. Model is given a table $x$ consisting of semantic fields and is tasked with generating a description $y_{1:T}$ of this data. Two example datasets are shown. Left: E2E, Right: WikiBio. }
\end{figure*}

\section{Experimental Setup}

\paragraph{Data and Metrics}
 We consider two standard neural generation benchmarks: E2E \citep{e2edata} and WikiBio \citep{wikibiodata} datasets, with examples shown in \cref{fig:data}. The E2E dataset contains approximately 50K examples with 8 distinct fields and 945 distinct word types; it contains multiple test references for one source table. We evaluate in terms of BLEU \citep{bleu}, NIST \citep{belz-reiter-nist}, ROUGE-L \citep{lin-2004-rouge}, CIDEr \citep{cider} and METEOR \citep{Lavie-meteor}, using the official scoring scripts\footnote{Official E2E evaluation scripts available at \url{https://github.com/tuetschek/e2e-metrics}}.
The WikiBio dataset contains approximately 700K examples, 6K distinct table field types, and 400K word types approximately; it contains one reference for one source table. We follow the metrics from \cite{wikibiodata} and evaluate the BLEU, NIST, and ROUGE-4 scores.

\paragraph{Architecture and Hyperparameters}
For all tasks, we use an encoder-decoder LSTM for the generative model. We follow recent state-of-the-art works in parametrizing our encoder, and we use copy attention and dual attention \citep{gu-etal-2016-incorporating, gulcehre-etal-2016-pointing,Liu-seq2seq}: full model architectures are given in the supplement.

The inference network scores are computed using a BiLSTM.  We compute the emission scores $\phi_{(e)}$ using span embeddings \citep{wang-chang-lstm-minus, kitaev-klein-lstm-minus, stern-etal-2017-lstm-minus}; transition scores $\phi_{(t)}$ by dot product between embedding vectors for the class labels; lengths $\phi_{(l)}$ is kept uniform, as in \newcite{template-generation}. Additional details are in the supplement.

At training time, we use a rate for alleviating posterior collapse in the ELBO: warm-up the ELBO objective by linearly annealing the coefficient on the term $\sum_{t=1}^{T} \log p_\theta(z_t \mid z_{<t}, y_{<t})$ and $\Entr[q_\phi(z\mid x,y)]$ from 0 to 1, as implemented in \citet{URNNG}. We use the REINFORCE algorithm to do Monte Carlo estimation of the stochastic gradient. We choose the control variate to be the mean of the samples \citep{pmlr-v48-mnihb16}.

At decoding time, we only use the generative model. We use beam search with length normalization to jointly generate both the control states and the sentences. To obtain controlled generation,  we observe the control states, and  apply constrained beam search to $p(y \mid  x,z)$.

\paragraph{Baselines}
\begin{table*}[]
\small
\begin{tabular}{lccccc}
\toprule
& \multicolumn{5}{c}{E2E}\\

         & BLEU  & NIST & ROUGE & CIDEr & MET \\
\midrule
& \multicolumn{5}{c}{validation}\\
\midrule
\exper{E2E-Bench*}  			& 69.25 & 8.48 & 72.6 & 2.40  & 47.0  \\
\exper{EncDec*}      			& 70.81 & 8.37 & 74.1  & 2.48 & 48.0   \\
\exper{NTemp}    			& 64.53 & 7.66 & 68.6 & 1.82  & 42.5  \\
\exper{NTemp+AR} 			& 67.70 & 7.98 & 69.5 & 2.29  & 43.1  \\
\Ours$^{0}$       	& 69.10  & 8.32 & 72.6 &  2.35 & 47.3 \\
\Ours$^{\infty}_{\texttt{}}$ 	& 69.36 & 8.36 & 71.3 &  2.29 & 46.4 \\
\Ours$_{\texttt{}}^\lambda$  & 72.93 & 8.63 & 75.5  & 2.54  & 48.4\\
\midrule
& \multicolumn{5}{c}{test}\\
\midrule
\exper{E2E-Bench*}   & 65.93 & 8.59 & 68.5 & 2.23  & 44.8  \\
\exper{Shen19*} & 68.60 & 8.73 &70.8 & 2.37 & 45.3  \\
\exper{EncDec*}      	& 66.34 & 8.55 & 68.0 & 2.18  & 44.3  \\
\exper{NTemp}    	& 55.17 & 7.14 & 65.7 & 1.70  & 41.9  \\
\exper{NTemp+AR} 	& 59.80 & 7.56 & 65.0 & 1.95  & 38.8  \\
\Ours$_{}^\lambda$      & 67.12 & 8.52 & 68.7  & 2.24  & 45.4\\
\bottomrule
\end{tabular}
\centering
\small \hspace*{0.5cm}
\begin{tabular}{lccc}
\toprule
& \multicolumn{3}{c}{WikiBio}\\
         & BLEU  & NIST & R-4 \\
\midrule
& \multicolumn{3}{c}{test}\\
\midrule
\exper{Lebret16}*   & 34.7 & 7.98 & 25.8 \\
\exper{Liu18(EncDec)}*   & 43.7 &   -  & 40.3\\
\exper{Liu18}(FieldGating)*   & 44.9 &   -  & 41.2\\
\exper{NTemp}         & 34.2 & 7.94 & 35.9\\
\exper{NTemp+AR}      & 34.8 & 7.59 & 38.6 \\
\Ours$^\lambda_{\text{one-to-one}}$          & 44.7 & 9.92& 43.3 \\
\Ours$^\lambda_{\text{one-to-many}}$ 		  & 44.2 & 9.59 & 41.5 \\
\bottomrule
\end{tabular}
\caption{\label{tb:e2e-scores} Automatic metrics for text generation. 
$*$ marks systems without learned control states.  
(Left) E2E. Comparison of systems from \newcite{dusek-jurcicek-2016-sequence,template-generation,shen-etal-2019-pragmatically}, our model and ablations. (Right) WikiBio. Comparison of \newcite{template-generation, Liu-seq2seq, wikibiodata}  and our full model. }
\end{table*}

For generation on E2E, we compare externally against 4 systems: 
\exper{E2E-Benchmark}  \citep{dusek-jurcicek-2016-sequence} is an encoder-decoder network followed by a reranker used as the shared task benchmark; \exper{NTemp}, a controllable neuralized hidden semi-Markov model; \exper{NTemp+AR}, the product of experts of both a NTemp model and an autoregressive LSTM network \cite{template-generation}; \exper{Shen19} \citep{shen-etal-2019-pragmatically} is an pragmatically informed model, which is the current state-of-the-art system on E2E dataset.

We  also compare internally with ablations of our system: 
\textbf{\exper{EncDec}} is a conditional model $p(y\mid x)$ trained without control states. 
\textbf{\exper{\Ours$^0$}} is posterior control model with no constraints. It uses structured encoder with the PR coefficient set to $0$. 
\textbf{\exper{\Ours$^\infty$}} is our model with hard constraints, which assumes fully-observed control states. These control states are obtained by mapping tokens with lexical overlap to their designated state; otherwise we map to a generic state. We train a seq2seq model $p(y,z\mid x)$ with full supervision of both control states and target text. 
Our main model is \textbf{\exper{\Ours$^\lambda$}}, which applies PR with coefficient given by hyperparameter $\lambda$.

For  WikiBio, we compare externally against 5 systems: \exper{NTemp} and \exper{NTemp+AR} as above; \exper{Lebret16} \cite{wikibiodata}, which uses copy attention and an NNLM; \exper{Liu18 (EncDec)}, which is our base encoder-decoder LSTM model, and \exper{Liu18} (Field Gating) which uses a field gating table encoder and a decoder with dual attention \cite{Liu-seq2seq}.  For internal comparison on WikiBio, we compare between the one-to-one and one-to-many constraints in \cref{sec:prNLG}. \textbf{\Ours$^\lambda_{\text{one-to-one}}$} applies the One-to-One posterior constraints (left of \cref{tb:pr-constraint}). \textbf{\Ours$^\lambda_{\text{one-to-many}}$} applies  the One-to-Many posterior constraints (right of \cref{tb:pr-constraint}).

\section{Experiments}
 
\cref{tb:e2e-scores} shows the main results for the E2E  and WikiBio, comparing to both standard neural models and controllable systems. On E2E (left), our posterior control model outperforms the neural benchmark system on all validation metrics and most of the test metrics. It also achieves results comparable or better than a specialized  encoder-decoder system. It has significantly better performance than the controllable NTemp and NTemp+AR in all metrics on both validation and test. This demonstrates that the PC model provides interpretable and controllable states without sacrificing any representation power or generation performance.

For internal comparison, having soft constraints on the posterior outperforms the system  \Ours$^{\infty}_{\texttt{}}$  (forced hard constraints) and \Ours$^{0}$  (no constraints). 
Anecdotally, we find that if two fields have the same value, then the hard coding system is often forced into the wrong decision. Similarly removing posterior regularization altogether leads to a slightly weaker performance than our controlled model. 

On the larger WikiBio dataset (right) our model also significantly outperforms both the controllable NTemp and NTemp+AR baselines in all three metrics. It gives improvements over \citet{Liu-seq2seq}'s strong encoder-decoder style model. 
The promising result from WikiBio dataset suggests that the method scales to larger datasets and the PR style works well in handling large field spaces. In addition, we find that dynamic constraints 
are feasible compared with static constraints (we believe this is  because the modeling burden on \Ours$^\lambda_{\text{one-to-many}}$ is heavier since it also needs to figure out the clustering).
Overall, the dynamic framework opens up the possibility of generalizing to work well with a wider set of constraints.

\section{Analysis}

\paragraph{Qualitative Analysis}
\cref{tb:wb-example} shows how control states (shown by different colors) are used in generated sentences. We use examples generated by the \Ours$^\lambda$ system on the WikiBio dataset. We obtain outputs by beam search over control states and words. The first block contains examples with relatively complete coverage by the semantically grounded control states, including name, birth date, death date, occupation and nationality. We note that when a control state is selected, the textual span covered by the control state tend to respect truthfulness by copying from the table. The second block shows a longer example that uses less of the source, but still remain truthful with respect to the table.

\cref{tb:control_example} (left) qualitatively demonstrates the multi-modality of output of the system on E2E dataset. We particularly note how the final system is trained to associate control states with field types. Here we fix the prior on $z$ to 8 different sequences of class labels shown in different colors, and do constrained beam search on the generative model by holding $z$ fixed, and decoding from the model $p_\theta(y\mid x,z)$.

\paragraph{Controllability}

\begin{table}[]
\small
\begin{tabular}{p{7.5cm}}

\toprule
\textsc{\Ours$^\lambda$}\\

\ratee{billy} \ratee{ruge} \otherr{-lrb-} \foodd{c.} \foodd{1885} \otherr{--} \areaa{1955} \otherr{-rrb-} \otherr{was} \otherr{an} \otherr{american} \pricee{film} \pricee{actor} \otherr{.} \\

\ratee{debra} \ratee{dene} \ratee{barnes} \otherr{is} \otherr{an} \pricee{associate} \pricee{professor} \pricee{of} \pricee{piano} \pricee{studies} \otherr{at} \restt{miss} \restt{america} \restt{1968} \otherr{.} \\

\ratee{shaalin} \ratee{zoya} \otherr{-lrb-} \otherr{born} \foodd{22} \foodd{february} \foodd{1997} \otherr{-rrb-} \otherr{is} \otherr{an} \eatt{indian} \pricee{actress} \otherr{.} \\

\ratee{carlos} \ratee{albert} \ratee{andrés} \otherr{-lrb-} \otherr{born} \foodd{february} \foodd{24} \foodd{,} \foodd{1978} \otherr{in} \nearr{madrid} \nearr{,} \nearr{spain} \otherr{-rrb-} \otherr{is} \otherr{a} \eatt{spanish} \otherr{sculptor} \otherr{.} \\
\bottomrule
\\
\toprule
Table ($x$): name[james horton]; birthdate[1850]; deathdate[none]; 
birthplace[boston, massachusetts];	
allegiance[united states of America];
branch[united states navy];  
rank[captain of the top];
awards[medal of honor] \\ \\
\textsc{Ref:} james horton -lrb- born 1850 -rrb- was a sailor serving in the united states navy who received the medal of honor for bravery .\\ \\
\textsc{\Ours$^\lambda$:} \ratee{james} \ratee{horton} \otherr{-lrb-} \otherr{born} \foodd{1850} \otherr{,} \otherr{date} \otherr{of} \otherr{death} \otherr{unknown} \otherr{-rrb-} \otherr{was} \otherr{a} \restt{united} \restt{states} \restt{navy} \otherr{sailor} \otherr{and} \otherr{a} \otherr{recipient} \otherr{of} \otherr{the} \restt{united} \restt{states} \otherr{military} \otherr{'s} \otherr{highest} \otherr{decoration} \otherr{,} \otherr{the} \ratee{medal} \ratee{of} \ratee{honor} \otherr{.} \\

\bottomrule
\end{tabular}
\caption{\label{tb:wb-example} Qualitative examples on WikiBio dataset.  (Top) Generated sentences control states highlighted. (Bottom) Full example of content selection with data table and reference. (Best viewed in color.)} 
\vspace{-2pt}
\end{table}

\begin{table*}[]
\hspace*{-1cm}
 \begin{minipage}{0.7\textwidth}
\centering
\small
\begin{tabular}{p{10cm}}
\toprule
Table ($x$): name[Clowns] eatType[coffee shop] food[English] \\ customerrating[5 out of 5]  area[riverside] near[Clare Hall]   \\
\midrule
(1) \restt{Clowns} \otherr{is} \otherr{a} \ratee{5} \otherr{star} \eatt{coffee} \eatt{shop} \otherr{located} \otherr{near} \nearr{Clare} \nearr{Hall} \otherr{.}   \\
(2) \restt{Clowns} \otherr{is} \otherr{a} \eatt{coffee} \eatt{shop} \otherr{that} \otherr{serves} \foodd{English} \otherr{food} \otherr{and} \otherr{is} \otherr{near} \nearr{Clare} \nearr{Hall} \otherr{.} \otherr{It} \otherr{is} \otherr{in}  \areaa{riverside}  \otherr{and} \otherr{has} \otherr{a} \ratee{5} \ratee{out} \ratee{of} \ratee{5} \otherr{customer} \otherr{rating} \otherr{.}   \\
(3) Near  \nearr{Clare} \nearr{Hall} in  Riverside  is  \eatt{coffee} \eatt{shop} \otherr{,} \restt{Clowns} \otherr{.} \otherr{It} \otherr{serves} \foodd{English} \otherr{food} \otherr{,}  \otherr{and}  \otherr{has} \otherr{received} \otherr{a} \otherr{customer} rating  of  \ratee{5} \ratee{out} \ratee{of} \ratee{5} \otherr{.}    \\
(4) Near  the  \areaa{riverside} ,  \nearr{Clare} \nearr{Hall} \otherr{is} \otherr{a} \eatt{coffee} \eatt{shop} \otherr{called} \restt{Clowns} \otherr{that} \otherr{serves}  \foodd{English} \otherr{food} \otherr{and} \otherr{has} \otherr{a} \otherr{customer} \otherr{rating} \otherr{of} \ratee{5} \otherr{-} \otherr{stars} \otherr{.}  \\
(5) Near  \nearr{Clare} \nearr{Hall} ,  \restt{Clowns} \eatt{coffee} \eatt{shop} \otherr{has} \otherr{a} \otherr{five} \otherr{star} \otherr{rating} \otherr{and} \foodd{English} \otherr{food} \otherr{.} \\
(6) \nearr{Clare} \nearr{Hall} \otherr{is} \otherr{a} \ratee{5} \otherr{star} \eatt{coffee} \eatt{shop} \otherr{near} \otherr{to} \restt{Clowns} \otherr{that} \otherr{serves} \otherr{British} \otherr{food} \otherr{.} \\
(7) \restt{Clowns} \eatt{coffee} \eatt{shop} \otherr{is} \otherr{near} \nearr{Clare} \nearr{Hall} \otherr{in} \otherr{Riverside} \otherr{.} \otherr{It} \otherr{serves} \foodd{English} \otherr{food} \otherr{and}  \otherr{has} \otherr{an} \otherr{excellent} \otherr{customer} \otherr{rating} \otherr{.} \\
(8) \ratee{5} \otherr{star} \otherr{rated} \otherr{restaurant} \otherr{,} \restt{Clowns} \eatt{coffee} \eatt{shop} \otherr{is} \otherr{located} \otherr{near} \nearr{Clare} \nearr{Hall} \otherr{.}  \\
\bottomrule
\end{tabular}
 \end{minipage} 
 \hspace{-1pt}
  \begin{minipage}{0.3\textwidth}
 \small
  \begin{tabular}{llll}
\toprule
Models    & Rec. $\downarrow$ & PPL $\downarrow$ & KL  \\
\midrule 
& \multicolumn{3}{c}{E2E} \\
\midrule
\Ours$^0$       & 1.81 & 3.74 & 19.8  \\
\Ours$^{\lambda}$         & 2.35 & 3.70 & 12.8 \\  
\midrule
& \multicolumn{3}{c}{WikiBio} \\
\midrule
\Ours$^0$      	& 2.57 & 3.82 & 10.69 \\ 
\Ours$^{\lambda}_{\text{one-to-one}}$  			& 2.45 & 4.07 & 10.19 \\ 
\Ours$^{\lambda}_{\text{one-to-many}}$ 		& 2.59 & 4.58 & 13.07  \\
\bottomrule
\end{tabular}

 \end{minipage}
\caption{\label{tb:control_example}  (Left) Example of controlled generation $p_\theta(y \mid x,z)$ on the source entity ``Clowns'' from E2E dataset. The color represents the class label of the token $z$. (Right) Metrics related to the generative model/inference network measured on both E2E and WikiBio. Rec. is reconstruction perplexity based on $\mathbb{E}_{q(z\mid x,y)}[\log p_\theta(y \mid, x,z)] $. PPL is the perplexity per token estimated by importance sampling.}
\end{table*}

Next we consider a quantitive experiment on model control. 
Assuming we have a mapping from control states to fields, ideally, at test time $z$ 
should use the right states from the source $x$.\footnote{On E2E dataset, we remove the binary table field, ``family friendly'' which is never expressed by lexical match.} Let ${\cal S} = \{(i, j, f): z_{i,j}=c, f \in x, \sigma(f) = c\}$ be the field states used by $z$. Define the field word overlap between $x$ and $y$ as, 
\[\#match = \sum_{(i,j, f)\in {\cal S}} \text{unigram-overlap}(y_{i:j}, x_f)\]
We can compute \textit{precision}, \textit{recall}, and \textit{coverage} under this metric,
$$ 
\frac{\#match}{\sum_{(i,j, f) \in \cal{S}} (j-i)}, \ \ \  \frac{\#match} { \sum_{f \in x} |x_f| }, \ \ \  \frac{|\cal{S}|}{ |{c : c  \in x}|}. 
$$ Under these metrics we see the following control metrics on the E2E dataset,

\begin{center}
\begin{tabular}{lcccccc}
\toprule
 & P  & R  & C & \\
\Ours$^{\infty}_{\texttt{}}$ & 0.996 & 0.895 & 0.833 &      \\ 
\Ours$^{\lambda}$ & 1.0   & 0.969  & 1.0 &  \\
 \bottomrule
\end{tabular}
\end{center}

\noindent
 The \Ours\ model with soft posterior constraints performs better than having hard constraints on all three metrics. Having $P=1$ means that the control states are a strong signal to copy from the table, and  $C=1$ means that control states learn to cover all table fields. On WikiBio, the model has a precision of $0.83$ on the, meaning that on average, when we generate a good control state, 83\% of the generated tokens will match the table content. Since only a fraction of the source table in WikiBio is used, recall and coverage are less applicable.

\paragraph{Distributional Metrics} 
Table~\ref{tb:control_example} (right) shows distributional metrics related to the optimization of the generative model and the inference network. The reconstruction perplexity, Rec. is much lower than the full perplexity,  PPL and the KL divergence between the variational posterior and the conditional prior is highly non-zero. These observations indicate that latent variables are being used in a non-trivial way by the generative model. It also suggests the variational model is not experiencing posterior collapse.

\paragraph{Limitations} Given the promise of PR as a technique for inducing control states, it is worth noting some of the current limitations to our specific application of the method.  Currently, we use simple rules which do not generalize well to paraphrase. Our weak supervision relies on direct overlap to align states and fails on aligning phrases like \texttt{less then 10 dollars} that are expressed as \texttt{cheap}. Additionally, while at test time, our method is comparable to a standard decoder model, it does require slightly longer to train due to both the dynamic program and the requirement to compute multiple samples.

\section{Conclusion}

This work introduces a method for controlling the output of a blackbox neural decoder model to follow weak supervision. The methodology utilizes posterior regularization within a structured variational framework. We show that this approach can induce a fully autoregressive neural model that is as expressive as standard neural decoders but also utilizes meaningful discrete control states. We show this decoder is effective for text generation while inducing meaningful discrete representations.  

Induction of grounded control states opens up many possible future directions for this work. These states can be used to provide integration with external rule-based systems such as hard constraints at inference time.  They also can be used to provide tools for
human-assisted generation. Another direction is to improve the sources of weak supervision and such as interactive new constraints provided by users. One could also explore alternative posterior constraints based on pre-trained models for summarization or paraphrase tasks to induce semantically grounded latent variables. Finally, it would be interesting to explore alternative training methods for these models, such as reducing reliance on hard sampling through better relaxations of structured models.

\section*{Acknowledgments} 
Thanks to Yoon Kim, Jambay Kinley, and Tristan Yang for ideas and discussion. AMR was supported by NSF CAREER 1845664 and IIS 1901030. XLL was supported by a Sony Research Award.  We thank the anonymous reviewers for helpful comments.\par

\bibliography{acl2020}
\bibliographystyle{acl_natbib}
\clearpage

\section*{Appendix}
The generative model is an LSTM with two layers with hidden dimension equals 500, input dimension equals 400, and dropout of 0.2. The inference network uses a one-layer Bi-LSTM with hidden size of 500 and input size of 400 to encode the sentence. We use large max segment length, $L=8$ (segmental for data-to-text) and $L=1$ (linear chain for POS induction) and 0.2 dropout in the inference network. The Bi-LSTM used for encoding the source table is has hidden dimension of $300$. 
Both the generative model and the inference network share word embeddings.

The batch size is $10$ for WikiBio and $20$ for PTB and E2E. 
The generative model and the inference network are optimized by Adam \citep{kingma2014adam} gradient clipping at $1$, with learning rate of $0.002$ and $0.001$ respectively. Parameters are all initialized from a standard Gaussian distribution. The learning rate decays by a factor of two for any epoch without improvement of loss function on validation set, and this decay condition is not triggered until the eighth epoch for sufficient training.  Training is done for max of 30 epochs and allows for early stopping.

 For data-to-text problem, we need to encode the data table. We encode the E2E source table by directly concatenating word embeddings and field embeddings and indices for each token, for example, if the word $w$ is the $i$th token from left and $j$th token from right under field type $f$, then we represent the token using a concatenation $[\text{emb}(w) \cdot \text{emb}(f) \cdot \text{emb}(i) \cdot \text{emb}(j)]$.  We encode the WikiBio table by passing a bidirectional-LSTM through the tokens in the table, where each token has similar embedding by concatenation as above. The encoding of the table is denoted as $c$. We use copy attention \citep{gu-etal-2016-incorporating, gulcehre-etal-2016-pointing} in the generative model, and the attention vector $\alpha$ at a time step is parametrized by the class label $z$ at that time step. Recall the contextual representation is $ \sum_{i} \alpha_i \cdot c_{i}$, where $\alpha_i = \Softmax(\score(h_t, c_i))$ and $\score(h_t, c_i) = (W_z(h_t) + b_z) \cdot (W_2(c_i) + b_2)$, the parametrization from $z$ happens during the feedforward network indexed by $z$.  For the WikiBio data, we use a dual attention mechanism described in \cite{Liu-seq2seq}, where the first attention is the same as above and the second attention uses a different encoder context $c_i'$, the $c_i'$ only looks at the concatenation of field type and field index, but not the field value itself, i.e. $[\text{emb}(f) \cdot \text{emb}(i) \cdot \text{emb}(j) ]$. Then the two attention forms two different sets of $\alpha_i$ and they are multiplied together and renormalized to form an attention. 

\end{document}